\documentclass[conference]{IEEEtran}
\IEEEoverridecommandlockouts
\usepackage{cite}
\usepackage{float}
\usepackage{caption}
\usepackage{multirow}
\usepackage{amsmath,amssymb,amsfonts}
\usepackage{algorithmic}
\usepackage{graphicx}
\usepackage{textcomp}
\usepackage{xcolor}
\usepackage{hyperref}

\def\BibTeX{{\rm B\kern-.05em{\sc i\kern-.025em b}\kern-.08em
    T\kern-.1667em\lower.7ex\hbox{E}\kern-.125emX}}
\begin{document}

\title{Detection of Micromobility Vehicles in Urban Traffic Videos\\
}

\author{
\IEEEauthorblockN{Khalil Sabri, Célia Djilali, Guillaume-Alexandre Bilodeau, Nicolas Saunier}
\IEEEauthorblockA{Polytechnique Montr\'eal\\Montr\'eal, Canada\\
\{khalil.sabri,celia.djilali,gabilodeau,nicolas.saunier\}@polymtl.ca }

\and
\IEEEauthorblockN{Wassim Bouachir}
  \IEEEauthorblockA{
  University of Qu\'ebec (T\'ELUQ)\\
    Montr\'eal, Canada\\
    wassim.bouachir@teluq.ca}
}

\maketitle

\begin{abstract}
Urban traffic environments present unique challenges for object detection, particularly with the increasing presence of micromobility vehicles like e-scooters and bikes. To address this object detection problem, this work introduces an adapted detection model that combines the accuracy and speed of single-frame object detection with the richer features offered by video object detection frameworks. This is done by applying aggregated feature maps from consecutive frames processed through motion flow to the YOLOX architecture. This fusion brings a temporal perspective to YOLOX detection abilities, allowing for a better understanding of urban mobility patterns and substantially improving detection reliability. Tested on a custom dataset curated for urban micromobility scenarios, our model showcases substantial improvement over existing state-of-the-art methods, demonstrating the need to consider spatio-temporal information for detecting such small and thin objects. Our approach enhances detection in challenging conditions, including occlusions, ensuring temporal consistency, and effectively mitigating motion blur.
\end{abstract}

\begin{IEEEkeywords}
 Urban Traffic, Micro-Mobility Detection, Object Detection, YOLO, Video Object Detection, Autonomous Vehicles, Urban Transportation Safety
\end{IEEEkeywords}

\section{Introduction}

In recent years, urban transportation has undergone a significant transformation with the emergence of micromobility solutions, in particular electric version of bikes, scooters, and skateboards. These modes of transport, renowned for their agility and environmental benefits, are reshaping short-distance commutes in bustling cities. However, the integration of these vehicles into the urban landscape is not without challenges. Their unique manoeuvrability and smaller footprints, while advantageous for users, complicate detection amidst heavy traffic, elevating safety risks.

The objective of this work is to design a robust micromobility vehicles (MMV) detection system, a critical component in ensuring the safety of both riders and surrounding traffic, particularly in the era of automated vehicles. As well, counting MMVs, their origins and destinations is an important component of efficient city planning. As micromobility solutions become increasingly prevalent in urban landscapes, the integration of such detection systems is paramount, not only for current traffic dynamics but also as a foundational technology for the seamless operation of future automated transportation systems.

This paper presents a novel detection model, FGFA-YOLOX, that synthesizes the strengths of image-based and video-based object detection methodologies. Our approach leverages the rapid and efficient image analysis capability of the YOLOX framework \cite{ge2021yolox}, integrating it with the temporal context consideration of video object detection systems (\cite{zhu2017deep,zhu2017flow,wu2019sequence}). This fusion aims to enhance detection consistency and accuracy in urban traffic scenarios, addressing the unique challenges posed by micromobility vehicles. This is done by applying aggregated feature maps from consecutive frames processed through motion flow to the YOLOX architecture. With this strategy, our method can benefit from the large number of readily available pre-trained models for YOLOX. Aggregation of feature maps than enhances the capabilities of the detector, thanks to the combination of several of object views, with some more informative than others.

To rigorously evaluate the performance of our proposed model against state-of-the-art (SOTA) methods, we constructed a new custom dataset focused on micromobility scenarios. Our findings indicate that our proposed model achieves superior performance compared to SOTA methods, demonstrating the need to consider spatio-temporal information for detecting such small and thin objects. Our contributions can be summarized as follows.

\begin{enumerate}
    \item We propose a novel video object detection architecture, FGFA-YOLOX, adapted for MMV and taking advantage from both the capabilities of single-frame object detection (accuracy and speed, several available pre-trained models) and the richer feature representation of video object detectors;
    \item We constructed a new dataset that we make publicly available for the evaluation of MMV object detectors, along with our source code that we provide to ensure reproducibility and reusability. Our model weights, code, and dataset are publicly available for further research and replication of our results at \url{https://github.com/sabrikhalil/Micro_Mobility_Detection}.
\end{enumerate}

\section{Related work}

Few works have considered micromobility vehicle (MMV) detection. The study by Apurv et al. \cite{apurv2021detection} introduced an approach for identifying e-scooter riders using a system comprised of two distinct modules. The first module employs a pre-trained YoloV3 model \cite{redmon2018yolov3} for effective initial detection of pedestrians. Following this, the second module comes into play, which involves expanding the bounding boxes around the detected individuals to encompass the e-scooter along with the rider. This expansion is essential to ensure the entire vehicle, including both the rider and the e-scooter, is captured. The process is further refined using a MobileNetV2 classifier \cite{sandler2018mobilenetv2}, trained on a specialized dataset, to distinguish whether a person is with or without an e-scooter. This two-module system permits to accurately identify e-scooter riders. Building on this previous work, Gilroy et al. \cite{gilroy2022scooter} extended the research for overcoming occlusion challenges in urban environments. They modify the YoloV3 architecture, tailoring it to detect partially visible e-scooter riders more reliably. This enhancement significantly improves the detection rates in scenarios where riders are obscured. 

In contrast to these two previous works that focus on detecting a single vehicle category, e-scooter, our research aims to include a variety of MMVs. We thus aim to establish a more versatile framework. Our goal is to enhance MMV detection capabilities to not only address occlusion challenges, but also to incorporate resilience against motion blur, a frequent issue in rapidly moving vehicles, and to ensure consistent detection across video frames. 

In single-frame object detection, two primary categories emerge: two-stage and one-stage models. The two-stage models, exemplified by R-FCN \cite{dai2016r}, adopt a sequential approach. Initially, region proposals are generated and subsequently they are classified into different object categories. This intricate process, while delivering precision, tends to be more computationally intensive, leading to slower inference times compared to one-stage models. One-stage models, like YOLO \cite{redmon2016you} and its iterations like YOLOX \cite{ge2021yolox}, streamline the detection process. YOLO considers object detection as a single regression problem, directly moving from image pixels to bounding box coordinates and class probabilities. Nevertheless, for all single frame approaches, their frame-by-frame nature might lead to detection inconsistency in video due to challenges like motion blur and occlusions, highlighting the need for models that maintain temporal consistency.

In the evolving landscape of video object detection, Deep Feature Flow (DFF) \cite{zhu2017deep} marked a significant progress. DFF emerged as a response to the computational demand of traditional methods. By innovatively employing keyframe feature extraction and utilizing optical flow, DFF transfers these features to adjacent non-key frames. This methodology not only accelerates the detection process but also alleviates the computational burden, making it a pivotal development for more efficient video object detection. Building on DFF emphasis on efficiency, Flow-guided Feature Aggregation (FGFA) \cite{zhu2017flow} took a step further, enhancing the quality of the features. FGFA aggregates and adaptively weights deformed feature maps from neighbouring frames. This process, aided by optical flow, improves accuracy by integrating relevant data across frames, addressing challenges like motion blur and occlusion. Similarly, Sequence Level Semantics Aggregation (SELSA) \cite{wu2019sequence} was proposed with a focus on the semantic relationships between objects across frames. SELSA analyzes and aggregates features based on semantic similarities, leading to more contextually aware and precise object detection.

\section{Methodology}

\subsection{Motivation}
\begin{figure*}[htbp]
    \centering
    \includegraphics[width=\textwidth]{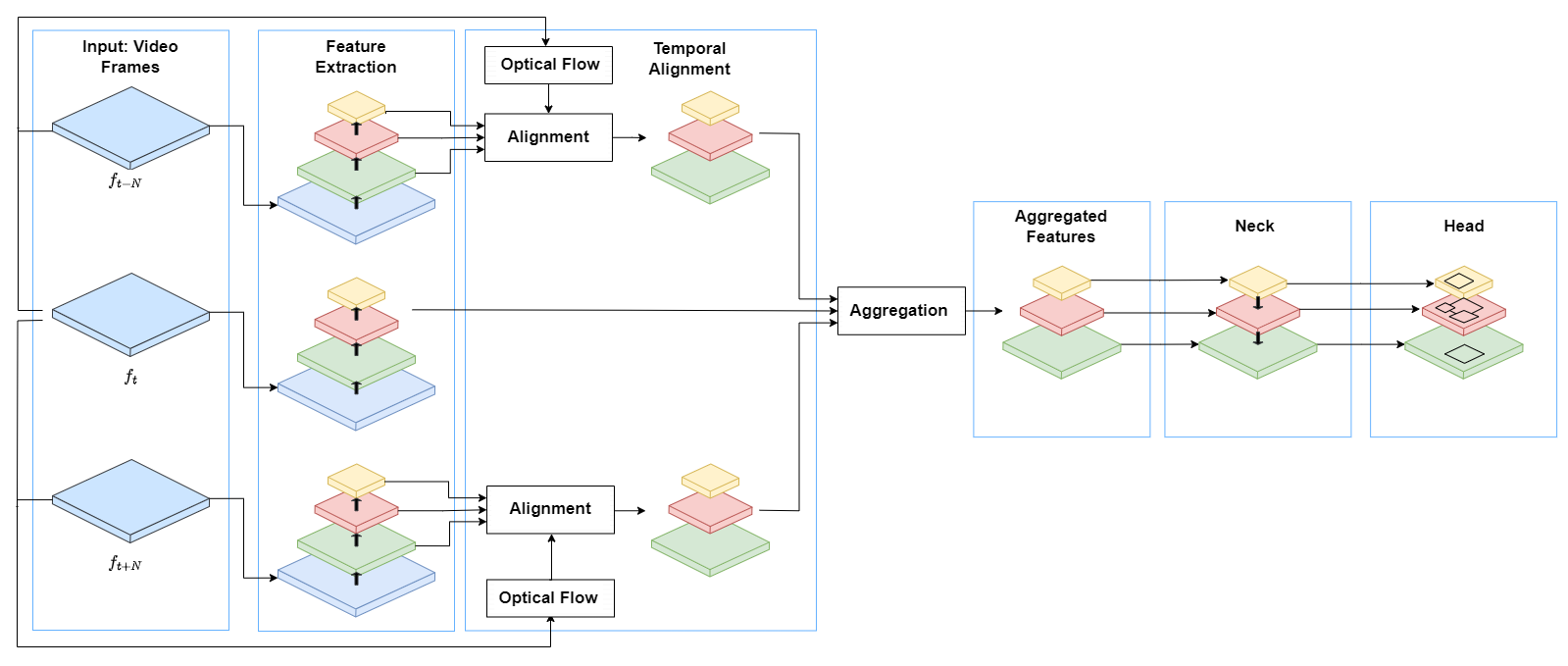}
    \caption{FGFA-YOLOX detection framework overview. Features from input frames are first extracted with the backbone of YOLOX. The optical flows of the current frame with past and future frames (neighbour frames) are also computed. For temporal aggregation, the motion-adjusted features of the neighbour frames are aggregated with those of the current frame. The aggregated features are then processes through the YOLOX architecture neck and head for detection.}
    \label{fig:micro_mobility}
\end{figure*}

While one-stage single-frame detection models like YOLOX \cite{redmon2016you} are efficient, they struggle with issues specific to videos, such as inconsistent detections over frames, motion blur, and occlusions. On the other hand, video object detection models that consider multiple frames provide better temporal consistency, but often lack the speed, the modern feature extraction strategies and the availabilty of pre-trained weights of one-stage models. To bridge this gap, we propose to combine the speed and accuracy of YOLOX \cite{ge2021yolox} with the temporal coherence of Flow-guided Feature Aggregation (FGFA) \cite{zhu2017flow}, aiming for a balanced solution in video object detection. We name our method FGFA-YOLOX. Figure~\ref{fig:micro_mobility} provides an overview of our proposed detection framework, showcasing the enhancement of the current frame feature map with motion-adjusted neighbour frames and subsequent processing through the YOLOX architecture neck and head for effective detection.

\subsection{Problem Statement}
Detecting micromobility vehicles (MMV) like e-scooters, bicycles, and skateboards in urban traffic involves challenges due to their size, movements, and the potential for occlusions. Given a sequence of video frames $F = \{f_{t-N}, \ldots, f_{t}, \ldots, f_{t+N}\}$, where $f_t$ is the current frame at time $t$ and $N$ indicates the number of contextual frames in the past and in the future, our goal is to accurately detect MMV in $f_t$. The goal is to find a function $D$ that, applied to the current frame $f_t$ and an aggregated feature map $\mathcal{G}$ derived from both $f_t$ and its contextual frames, yields detections defined by bounding boxes $B$, class labels $L$, and confidence scores $S$:

\begin{equation}
    D(f_t, \mathcal{G}) \rightarrow \{(B, L, S)\}.
\end{equation}

\subsection{Integrated Detection Model Framework}

\noindent\textbf{Feature Extraction:}
As shown in Figure~\ref{fig:micro_mobility}, the first step of our approach is to perform feature extraction. The CSPDarknet backbone $\mathcal{B}$ extracts spatial features $\mathcal{S}_t$ from the current frame $f_t$ and $\mathcal{S}_{\text{context}}$ from the neighbouring contextual frames, defined as:

\begin{equation}
    \mathcal{S}_t = \mathcal{B}(f_t),
\end{equation}
and 
\begin{equation}
\mathcal{S}_{\text{context}} = \bigcup_{i=t-N, i \neq t}^{t+N} \mathcal{B}(f_i).
\end{equation}

The CSPDarknet backbone, selected for feature extraction in our model, was shown to be very efficient and capable by Wang et al. \cite{wang2020cspnet}. This architecture is efficient in optimizing gradient flow and reducing computational load, making it ideal for detecting objects of varying sizes, particularly micromobility vehicles (MMV) in urban environments. For an input image, this backbone produces feature maps with dimensions $H \times W \times C$, where $H$ and $W$ represent the height and width of the reduced spatial resolution of the output feature maps, and $C$ symbolizes the channel depth, which is $C_1$, $C_2$, or $C_3$ for each scale, adapting to the pyramid feature network requirements. This downsampling is essential for processing high-resolution inputs efficiently while retaining significant spatial features necessary for accurate object detection.

\noindent\textbf{Temporal Alignment and Aggregation:}
As shown in Figure~\ref{fig:micro_mobility}, the second step is feature aggregation. A motion estimation function $\mathcal{M}$ aligns context frame features $\mathcal{S}_{\text{context}}$ with the current frame feature $\mathcal{S}_t$. This alignment ensures that the features are spatially coherent and results in features:

\begin{equation}
    \mathcal{S}_{\text{aligned}} = \mathcal{M}(\mathcal{S}_{\text{context}}, \mathcal{S}_t).
\end{equation}

FlowNetSimple \cite{dosovitskiy2015flownet} is used to implement the motion estimation function $\mathcal{M}$ to align feature maps from context frames to the current frame feature map. This process aligns $\mathcal{S}_{context}$, with dimensions $H \times W \times C$, to the spatial configuration of $\mathcal{S}_t$ to closely mirror the current frame feature map in both spatial and channel dimensions, effectively capturing the estimated features at time $t$. This alignment enables a more accurate representation of the scene at the current moment, laying the groundwork for subsequent feature aggregation to further enhance detection capabilities in evolving urban environments.

These aligned features are aggregated with $\mathcal{S}_t$ via the aggregation function $\mathcal{A}$, forming an enriched feature map $\mathcal{G}$ that captures both spatial and temporal characteristics:

\begin{equation}
    \mathcal{G} = \mathcal{A}(\mathcal{S}_{\text{aligned}}, \mathcal{S}_t).
\end{equation}

In our method, the aggregation function $\mathcal{A}$ employs 
concatenation and convolutions for merging $\mathcal{S}_{\text{aligned}}$ and $\mathcal{S}_t$. This operation is realized through a concatenation step, formulated as:

\begin{equation}
    \mathcal{S}_{\text{stacked}} = \text{Concat}(\mathcal{S}_{\text{aligned}}, \mathcal{S}_t),
\end{equation}

\noindent that effectively doubles the channel size of the input features, preparing them for the subsequent convolution process. The convolution, employing a $3 \times 3$ kernel, is designed to integrate and refine the concatenated features, resulting in:

\begin{equation}
    \mathcal{G} = \text{Conv}_{3 \times 3}(\mathcal{S}_{\text{stacked}}),
\end{equation}

\noindent where $\mathcal{G}$ is the output feature map with dimensions $H \times W \times C$, maintaining the original spatial dimensions while encapsulating an enriched representation of both spatial and temporal information.

\noindent\textbf{Detection and Classification:}
Finally, as shown in Figure~\ref{fig:micro_mobility}, detection and classification is performed through the neck and head. After aggregating features to obtain $\mathcal{G}$, the YOLOXPAFPN neck $\mathcal{N}$ is employed. The 'PAFPN' in YOLOXPAFPN denotes Path Aggregation Feature Pyramid Network, which enhances multi-scale feature fusion by leveraging the Path Aggregation Network (PAN) for efficient optimization \cite{liu2018path}. This configuration refines the feature processing across scales, optimally preparing them for the detection task. The YOLOX head \(\mathcal{H}\) processes the enriched feature map to detect Micromobility Vehicles (MMV) within the current frame \(f_t\). This sequential application of the neck and head on \(\mathcal{G}\) is given by:

\begin{equation}\label{eq:bls}
    (B, L, S) = D(f_t, \mathcal{G}) = \mathcal{H} \circ \mathcal{N}(\mathcal{G}).
\end{equation}

In more details, our method employs a feature pyramid network (FPN) \cite{lin2017feature} in the form of YOLOXPAFPN for integrating the multi-scale feature maps. This approach ensures good detection across various object sizes, crucial for smaller targets requiring high-resolution recognition. The process involves merging aggregated feature maps $\mathcal{G}_{C1}$, $\mathcal{G}_{C2}$, and $\mathcal{G}_{C3}$, each indicative of a unique scale within the detection framework:

\begin{equation}
    \mathcal{N(G)} = \text{FPN}(\mathcal{G}_{C1}, \mathcal{G}_{C2}, \mathcal{G}_{C3})
\end{equation}

This integration yields $\mathcal{N(G)}$, a composite feature map of $H \times W \times C$, standardizing the output to facilitate precise, scale-invariant object detection. Through this pyramid approach, the model effectively consolidates spatial and scale data, enhancing detection accuracy in varied urban environments.


The YOLOX head, $\mathcal{H}$, which is the concluding component in our detection framework, utilizes the feature map \(\mathcal{N(G)}\) to delineate to location of objects and classify them. It conducts a detailed analysis on \(\mathcal{N(G)}\), outputting a set of detections $\mathcal{D}$, each defined by a bounding box $B$, a class label $L$, and a confidence score $S$ as presented in equation~\ref{eq:bls}.


This operation is key to identifying and classifying MMV accurately, incorporating confidence scores \(S\) to gauge the reliability of each detection. By handling \(\mathcal{N}(\mathcal{G})\) from the YOLOXPAFPN neck, the head ensures detailed detection across varying object sizes, essential for monitoring dynamic urban scenes. This approach maintains consistency with our problem statement, underlining the necessity of combining current frame analysis with contextual features for robust and accurate MMV detection.

\section{Experiments}

This section outlines the experimental setup and the evaluation of our proposed model, including the dataset used, evaluation metrics, implementation details,  and comparison with state-of-the-art (SOTA) models.

\subsection{Dataset}

\subsubsection{Dataset Collection and Construction}
A custom dataset, named PolyMMV, was developed to address the detection of MMV, given the scarcity of suitable existing datasets. Sourced from a variety of online public video hosting platforms, this dataset aims to mirror the diversity of real-world urban micromobility, covering bicycles, skateboards, and electric scooters as primary classes. 

\begin{figure*}[h]
    \centering
    \includegraphics[width=120mm]{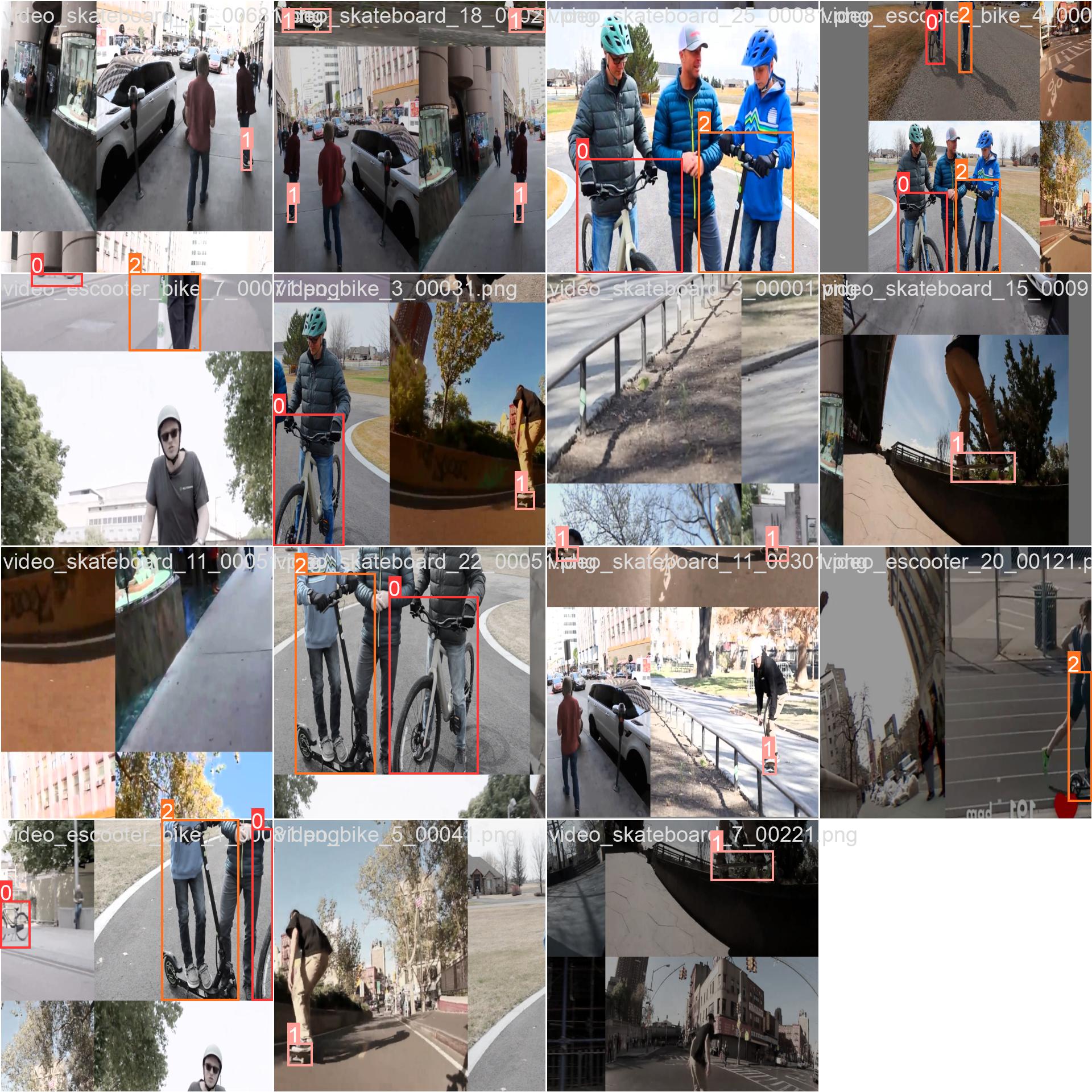}
    \caption{Examples of annotated images in PolyMMV. Class 0: Bicycles, Class 1: Skateboards, Class 2: Electric Scooters}
    \label{fig: train_batch1}
\end{figure*}

For research reproducibility and further exploration, the dataset, including annotations in YOLO, COCO, and VOC formats, is available at our GitHub repository: \url{https://github.com/sabrikhalil/Micro_Mobility_Detection}. This initiative supports the advancement of urban MMV detection research. 
Figure~\ref{fig: train_batch1} presents samples of annotated images from the training set. These examples, randomly chosen and representing various classes, showcase the real-world diversity in object sizes, positions, lighting conditions, and occasional occlusions. Some images also feature objects from multiple classes, adding to the dataset complexity.

\subsubsection{Annotation Process and Dataset Characteristics}
\begin{figure}[H]
    \centering
    \includegraphics[width=70mm,scale=1]{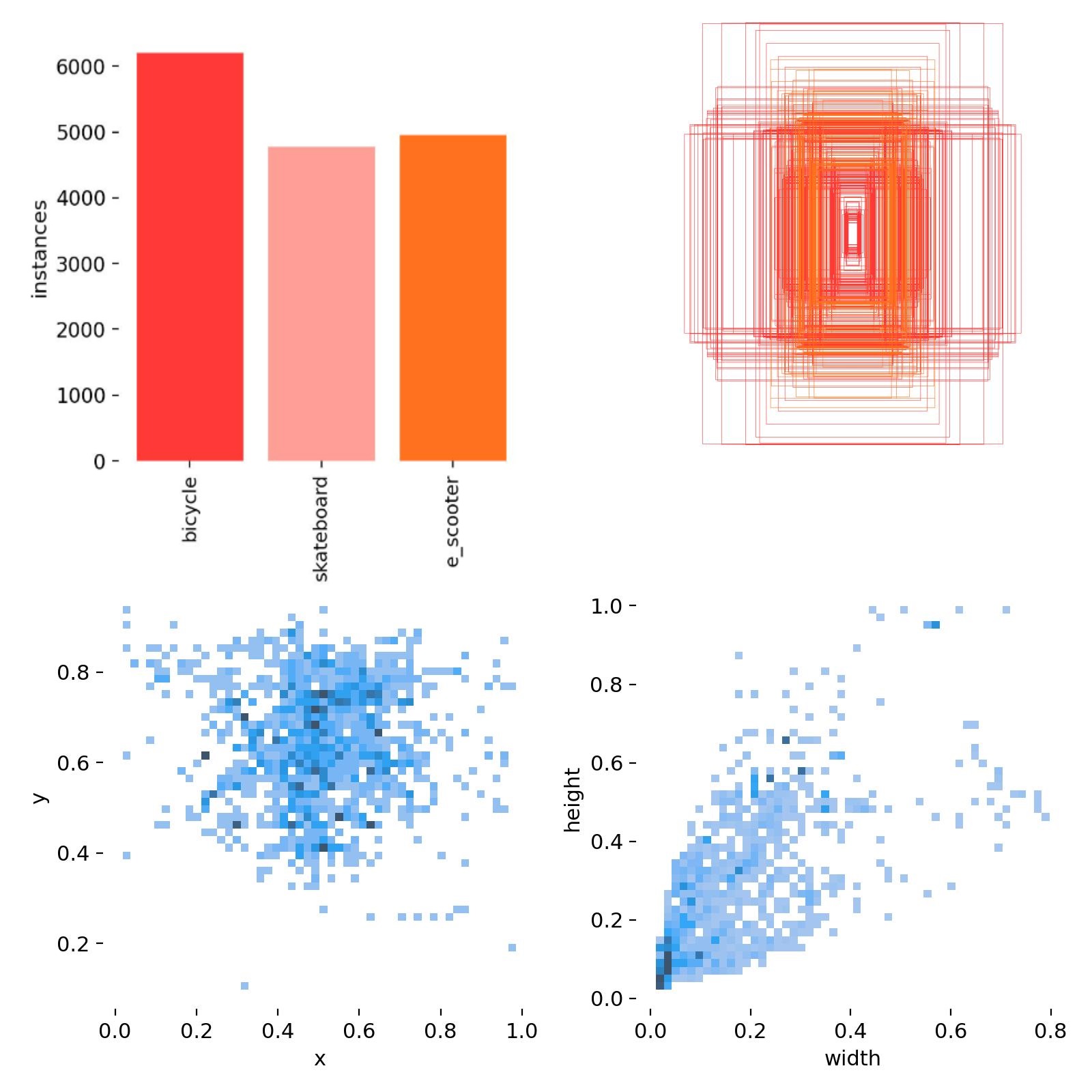}
    \caption{Characteristics of the training dataset of PolyMMV. Top left: the number of instances distribution of bicycles, skateboards, and e-scooters, top right:  illustration of the size distribution of bounding boxes, bottom left and right depict the scatter plots of normalized bounding box positions and sizes, respectively.}
    \label{fig:labels}
\end{figure}

The videos were annotated using the Computer Vision Annotation Tool (CVAT). The annotation process involved labelling bicycles, skateboards, and electric scooters with bounding boxes. Figure~\ref{fig:labels} shows the characteristics of our dataset. The training set contains 80 videos, and the test set comprises 25 videos. As seen in Figure~\ref{fig:labels}, the training data includes around 6000 bounding boxes for bicycles, and approximately 5000 each for electric scooters and skateboards. This balanced distribution facilitate the training of the model across different MMV classes. The diverse sizes and positions of the bounding boxes, representative of real-world scenarios, are also visible in the figure.

In contrast, the test set is designed to evaluate the model generalization capabilities and includes 4000 instances of bicycles, 2500 skateboards, and 2000 electric scooters.

\subsection{Evaluation Metrics}
For the evaluation of our model, we used the mean Average Precision (mAP) and mAP@50 metrics. The mAP provides an overall effectiveness of the model by averaging the precision across different recall levels and object categories, and mAP@50 specifically considers detections as correct if they have an Intersection over Union (IoU) of more than 50\% with the ground-truth. AP is similar to the mAP, but used for each individual object class. These metrics are particularly relevant in object detection tasks to measure the accuracy of the model in identifying and localizing objects correctly. Our evaluation metrics align with the standards set by the COCO evaluation protocols, which are widely recognized for their comprehensiveness in assessing object detection performance \cite{lin2014microsoft}.

\subsection{Implementation Details and Model Training}
Our models were implemented using Python and the PyTorch framework, alongside mmtracking \cite{mmtrack2020} for video object detection and OpenCV \cite{opencv_library} for video data handling. We integrated YOLOX into the FGFA framework, utilizing components that were pre-trained on specific datasets to enhance their robustness and effectiveness. The YOLOX backbone used for object detection, was pre-trained on the COCO dataset \cite{lin2014microsoft}. Its configuration, with a deepen factor of 1.33 and a widen factor of 1.25, effectively enhances the network ability to capture intricate details without excessive computational demands. This is crucial for real-time video analysis. The input image size is $640 \times 640$, giving values $C_1=320$, $C_2=640$ and $C_3= 1280$. The FlowNetSimple \cite{dosovitskiy2015flownet,ilg2017flownet} module is used for motion estimation and was pre-trained on the Flying Chairs dataset \cite{dosovitskiy2015flownet}, as described in the original FlowNet study.

Our FGFA-YOLOX model and the DFF-YOLOX variant underwent fine-tuning from COCO pre-trained weights using an SGD optimizer with an initial learning rate of 0.0001, momentum of 0.9, and weight decay of 0.00001 over three epochs, incorporating a warm-up phase during the first 500 iterations. Similarly, RFCN-based methods (DFF-RFCN, FGFA-RFCN, SELSA-RFCN) were initially trained with an R-FCN detector on COCO video data, with subsequent fine-tuning for our MMV detection task that adjusted the learning rate to 0.01, alongside comparable momentum and weight decay settings. For YOLOv8, originally designed for single-image object detection and also pre-trained on the COCO dataset, we adapted it for video detection by sampling at 10 frames per second to avoid overfitting from frame redundancy and utilized dropout techniques to improve generalization. This multifaceted approach to training, including specific learning rate adjustments and dataset-oriented refinements for each model, was crucial in enhancing detection capabilities across diverse urban environments.

\subsection{Comparison with State-of-the-Art Object detectors}

\begin{table*}[] 
    \centering
    \begin{tabular}
    {|l|c|c|c|c|c|c|}
        \hline
        Models & mAP  & mAP@50 & AP-bicycle & AP-skateboard & AP-scooter  & Inference Time per Frame \\
        \hline
        \hline
        DFF - RFCN \cite{zhu2017deep} & 27.9 & 57.6 & 33.8 & 8.2 & 41.7 & 41.4 \\
        FGFA - RFCN  \cite{zhu2017flow} & 31.2 & 61.1 & 38.9 & 10.0 & 44.7 & 418.1 \\
        SELSA - RFCN  \cite{wu2019sequence} & 31.0 & 62.4 & 36.6 & 11.6 & 44.7 & 317.0 \\
        YOLOv8 \cite{yolov8_ultralytics} & 34.5 & 64.2 & 39.2 & 17.7 & 46.7 & \textbf{34.2} \\
        \textbf{FGFA - YOLOX (ours)} & \textbf{38.6} & \textbf{69.4} & \textbf{45.0} & \textbf{23.2} & \textbf{47.6} & 329.7 \\
        \hline
    \end{tabular}
    \caption{Comparison of model performances and inference times on PolyMMV dataset. mAP and AP are in percentage. Inference time are in milliseconds. Best scores in \textbf{bold} font.}
    \label{table: map-time}
\end{table*}

Table \ref{table: map-time} presents a performance comparison between our proposed detector FGFA-YOLOX and various SOTA object detection models, highlighting its superior performance, achieving the highest scores in mAP and mAP@50. It particularly improves in detecting skateboards, a challenging category due to their smaller size, frequent occlusion by the rider's feet, and motion blur as evidenced in Figure \ref{fig:combined_scenarios}B. Bicycles also saw a considerable increase in detection accuracy, benefiting from the model's robust feature extraction and aggregation capabilities. In contrast, e-scooters, which tend to move slower and exhibit more consistent movement patterns, showed less improvement. This variance underscores the model's adaptability to different object characteristics within urban traffic scenarios.

\begin{table}[htbp]
    \centering
    \caption{Confusion matrix for the detection of micromobility vehicles on PolyMMV, assuming an IoU threshold of 0.5.}
    \label{table:confusion_matrix}
    \begin{tabular}{|c|c|c|c|c|}
        \hline
        & \multicolumn{4}{c|}{\textbf{Predicted Class}} \\ \cline{2-5}
        \textbf{Actual Class} & \textbf{Bicycle} & \textbf{Skateboard} & \textbf{E-scooter} & \textbf{Background}\\
        \hline
        \textbf{Bicycle} & 0.83 & 0 & 0 & 0.17\\
        \hline
        \textbf{Skateboard} & 0 & 0.44 & 0 & 0.56\\
        \hline
        \textbf{E-scooter} & 0.02 & 0 & 0.66 & 0.32\\
        \hline
        \textbf{Background} & 0.57 & 0.26 & 0.17 & 0\\
        \hline
    \end{tabular}
\end{table}

The confusion matrix for our FGFA-YOLOX confirms our previous observations. The model excels in separating classes by learning from several frames. Specifically, 44\% of skateboards are correctly detected while this number is 31\% for FGFA-RFCN and 35\% for YOLOv8 \cite{yolov8_ultralytics}, representing a significant improvement. However, there are areas for improvement, particularly in fine-tuning IoU thresholds to better balance the detection of micromobility vehicles against complex urban backgrounds. This adjustment is aimed at reducing the higher incidence of false positives observed, enhancing the model's precision. Further diversifying our training dataset will also support this goal, enabling the model to more accurately recognize electric scooters, bicycles, and skateboards in a variety of urban scenarios.

Table \ref{table: map-time} also illustrates the inference times of various detection models. Our FGFA-YOLOX model notably achieves faster inference times compared to FGFA-RFCN, but shows that using several frame for detecting is more costly than for example YOLOv8.

\subsection{Model Performance in Various Scenarios}

\begin{figure*}[]
    \centering
    \includegraphics[width=\textwidth]{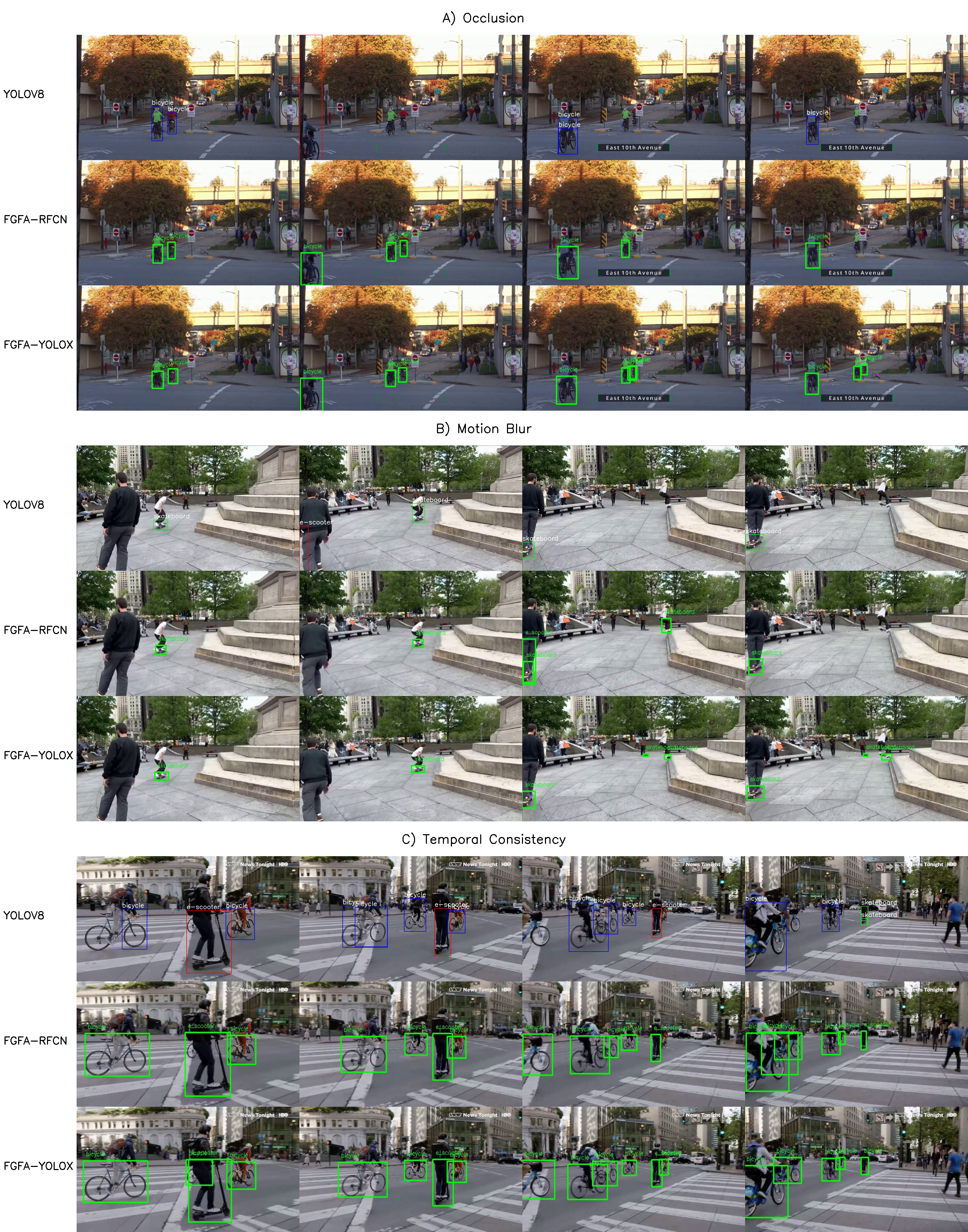}
    \caption{Comparative analysis of model performance in various scenarios. A) Occlusion, B) Motion blur, and C) Temporal consistency.}
    \label{fig:combined_scenarios}
\end{figure*}

To illustrate the performance of FGFA-YOLOX, we will discuss in the following sample qualitative results in three complex urban traffic scenarios: occlusion, motion blur, and temporal consistency. The figure \ref{fig:combined_scenarios} below contrasts our model performance with that of single-frame detection models like YOLOV8 and video-based models like FGFA-RFCN, showcasing our approach robustness in handling these challenges.

\textbf{Occlusion Handling:} In dense urban environments, vehicles often become partially occluded. Our model excels in such scenarios, effectively detecting MMVs despite occlusions. As depicted in figure \ref{fig:combined_scenarios}A, our model successfully identifies vehicles that are partially hidden, a task where the YOLOV8 model struggles due to its reliance on single-frame analysis.

\textbf{Motion Blur:} Figure \ref{fig:combined_scenarios}B illustrates our FGFA-YOLOX ability to handle motion blur caused by rapid movements, which is a significant challenge for single-frame models, like YOLOV8, that do not utilize past frames for continuous object identification and can be affected by motion blur. Unlike YOLOV8, our model mitigates the effects of motion blur, ensuring more detection of fast-moving objects, particularly skateboards.

\textbf{Improved Temporal Consistency:} Figure \ref{fig:combined_scenarios}C demonstrates that our model enhances temporal consistency. While traditional video object detection models may show variability in detections across frames, our model maintains stable and accurate detection, important for real-time monitoring and autonomous navigation in urban traffic.

The combined strengths of our model, merging the precision of single-frame detection with the comprehensive context of video object detection, offer a significant advancement in addressing the diverse challenges of urban traffic conditions for MMV detection.

\section{Conclusion}

This paper introduces a novel detection model that leverages both single-frame precision and video object detection capabilities for accurately identifying MMV in urban environments. Our approach, validated on a new MMV dataset, demonstrates significant improvements over existing methods by incorporating spatio-temporal information for enhanced detection performance.

Our research opens avenues for exploring attention mechanisms and innovative architectures for feature aggregation. Integrating attention mechanisms into the model could refine its focus on relevant features, improving occlusion handling and detection of small or partially visible vehicles.

\bibliographystyle{plain}
\bibliography{References}

\end{document}